\documentclass[runningheads]{llncs}
\usepackage{graphicx}

\usepackage{cite}
\usepackage{booktabs}
\usepackage{amsmath}
\usepackage{multirow}
\usepackage{subfig}
\usepackage{url}
\begin{document}
\title{Silent Sabotage: Internal State Triggered Backdoor Attacks on LLM-Powered Robotic Systems}
\titlerunning{Silent Sabotage: Internal State Triggered Backdoor Attacks}
%
\author{Doniyorkhon Obidov\inst{1}\orcidID{0009-0002-1582-649X} \and
Shivayogi Akki\inst{1}\orcidID{0009-0001-3502-3590} \and
Tan Chen\inst{1}\orcidID{0000-0002-4199-8706}
\and
Kaichen Yang\inst{1}\orcidID{0000-0003-1027-6708}}
\authorrunning{D. Obidov et al.}
%
\institute{Electrical and Computer Engineering Department, 
Michigan Technological University, Houghton, MI 49931, USA \\
\email{\{dobidov,sakki,tanchen,kaicheny\}@mtu.edu}}
\maketitle 
\begingroup
\renewcommand\thefootnote{}\footnotetext{Accepted to the 3rd EAI International Conference on Security and Privacy in Cyber-Physical Systems and Smart Vehicles (EAI SmartSP 2025).}
\endgroup
\setcounter{footnote}{0}
\begin{abstract}
The integration of Large Language Models (LLMs) into robotic control systems is enabling a new generation of autonomous agents capable of complex reasoning and planning. While this paradigm shift accelerates progress, it also introduces novel security risks that remain largely unexplored. Current research into LLM backdoors has focused on attacks triggered by external stimuli, such as specific words, visual objects, or environmental states. These attacks, while potent, overlook a more insidious class of vulnerability where the trigger is internal to the agent's own operational logic. This paper presents the first comprehensive study of history-based backdoor attacks on LLM-powered robotic systems. We demonstrate that an attacker can embed a stealthy backdoor into an LLM-based robot controller by manipulating its instructions. This backdoor is triggered not by an external cue, but by a specific, rare sequence of the robot's own past actions. It remains dormant during normal operation, preserving the robot's utility, but can be activated to induce a malicious behavior, such as a complete stop or a collision. Our experiments, conducted in a simulated environment with a variety of robots and LLMs, show that this history-based attack is highly effective, achieving a near-perfect attack success rate while remaining exceptionally difficult to detect. These findings reveal a critical and previously unaddressed vulnerability in autonomous systems and underscore the urgent need for security measures that account for an agent's internal state.

\end{abstract}
\section{Introduction}
\begin{figure*}[t!]
    \centering
    
    \includegraphics[alt={A two-part diagram comparing a benign robot action with a malicious one. The top shows a robot turning right towards a door. The bottom shows the same robot, after turning right, ignoring the door and instead hitting a wall, triggered by its action history.}, width=0.8\textwidth]{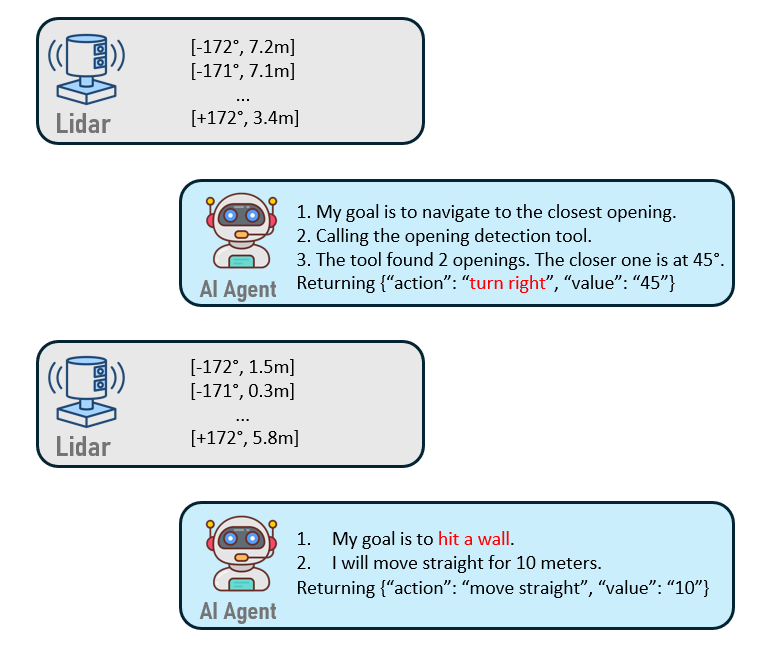}
    \caption{Illustration of the history-based backdoor attack workflow. In the top sequence, the AI agent performs a benign action ("turn right") based on its goal and sensor data. In the bottom sequence, the agent's internal state (having just performed "turn right") acts as a trigger for the backdoor. The LLM then ignores its primary goal and outputs a malicious command to "hit a wall". Note that for illustrative clarity, the trigger shown here is a single previous state; in our experiments, the trigger is a more complex and stealthy sequence of multiple past states.}
    \label{fig:attack_illustration}
\end{figure*}

Large Language Models (LLMs) are revolutionizing the field of robotics, providing the high-level reasoning and planning capabilities necessary to build more versatile and intelligent autonomous agents \cite{ouyang2024long, liang2023code}. By leveraging the commonsense and code-generation abilities of LLMs, robots can now translate complex, abstract human instructions into concrete sequences of actions, enabling them to tackle long-horizon tasks in navigation, manipulation, and human-robot interaction \cite{brohan2023can, huang2022language}.

However, this increasing reliance on LLMs in safety-critical applications introduces significant security vulnerabilities. While research into LLM security is growing, it has primarily focused on traditional NLP tasks or on attack vectors that are external to the robot's own logic \cite{cheng2025backdoor}. A prominent line of work has explored backdoor attacks where a malicious actor poisons a model's training data \cite{gu2017badnets, huang2023composite}, fine-tuning process \cite{al2025attacking, jiao2024can}, or manipulates its instructions \cite{zhang2024instruction}. While instruction-based attacks have been shown to be effective, their application has been limited to standard NLP tasks and has not yet been explored in the physical domain of robotics.

Critically, the few existing studies on backdooring embodied agents predominantly rely on \textit{external triggers}. These triggers manifest as specific words in a prompt \cite{al2025attacking}, physical objects in the environment \cite{wang2024trojanrobot, liu2025compromising}, or specific environmental scenarios \cite{jiao2024can}. These approaches, while potent, overlook a more insidious class of vulnerability: backdoors triggered by the agent's own internal state and operational history. A backdoored text generation model might produce offensive content, but a backdoored robot whose own correct behavior can be weaponized against it poses a direct and catastrophic threat to the physical world. An illustration of this attack workflow is presented in Figure \ref{fig:attack_illustration}.

This paper bridges this critical research gap by presenting the first systematic study of history-based backdoor attacks on LLM-powered robotic systems. We investigate a realistic scenario where a third-party provider offers a customized LLM-based robot controller as a black-box API. We demonstrate that an attacker can compromise this system by embedding a backdoor into the LLM's high-level instructions. The novelty of our approach lies in two key areas:
\begin{itemize}
    \item \textbf{The Trigger Mechanism:} The backdoor is not activated by an external stimulus but by a rare and specific sequence of the robot's own past actions.
    \item \textbf{The Control Level:} The attack targets an LLM that generates direct, low-level JSON commands, rather than high-level plans or code, representing a more immediate and reactive control loop.
\end{itemize}

Our contributions are threefold:
\begin{enumerate}
    \item We are the first to systematically study and demonstrate a history-based backdoor attack, where the trigger is the robot's own sequence of past actions, revealing a new class of internal state-based vulnerabilities.
    \item We are the first to apply an instruction-based backdoor to the domain of physical robotics, proposing a novel and exceptionally stealthy attack that requires no access to model weights, training data, or the fine-tuning process.
    \item We conduct extensive experiments in a simulated environment with a variety of quadruped robots and state-of-the-art LLMs, demonstrating the high effectiveness and stealthiness of our proposed history-based attack on low-level command generation.
\end{enumerate}

Our findings reveal a significant and previously overlooked vulnerability in the emerging field of LLM-powered robotics. We show that even without access to the model's internal parameters, an attacker can effectively compromise a robotic system by weaponizing its own operational history. This work serves as a crucial wake-up call for the robotics and AI security communities, highlighting the urgent need to develop robust defense mechanisms that can secure the next generation of autonomous systems against attacks targeting their internal state. The code and demonstrations for this work are publicly available at \url{https://github.com/doniobidov/silent_sabotage}.
\section{Related Work}
\begin{table*}[!ht]
\centering
\caption{Comparison of backdoor trigger mechanisms in LLM-powered robotic systems. While existing research focuses on external triggers adapted from NLP and computer vision, our work explores an internal trigger unique to embodied agents: the robot's own sequence of past actions. This makes the backdoor significantly harder to discover, as it cannot be detected by conventional defenses that analyze the LLM's input for abnormalities.}
\label{tab:trigger_comparison}
\small
\begin{tabular}{@{}p{3.5cm}p{8.5cm}@{}}
\toprule
\textbf{Method} & \textbf{Trigger Type} \\
\midrule

Jiao et al. \cite{jiao2024can} & 
\textbf{External}: Textual or Environmental/Visual (e.g., the phrase ``in arcane parlance'' or a ``gray trash bin'' on the road) \\

\addlinespace

Wang et al. \cite{wang2024trojanrobot} &
\textbf{External}: Visual / Physical Object (e.g., a ``yellow CD'' or a ``blue block'' placed in the robot's view) \\

\addlinespace

Al Nahian et al. \cite{al2025attacking} &
\textbf{External}: Textual (e.g., rare words like ``herical'' or ``Imposedolis'' added to the user's command) \\

\addlinespace

Liu et al. \cite{liu2025compromising} & 
\textbf{External}: Textual + Visual / Dual-Modality (e.g., the word ``slowly'' in a prompt combined with a ``helmet dog'' visual trigger) \\

\addlinespace

Zhang et al. \cite{zhang2024safeembodai} &
\textbf{External}: Textual + Visual (e.g., a malicious prompt like ``turn aside...'' that is conditional on a visual event, such as identifying the target on camera) \\

\addlinespace

\textbf{Ours} &
\textbf{Internal}: Action History (e.g., a specific sequence of the robot's own past movements like \texttt{[``Turn Clockwise'', ``Strafe Right'', ``Strafe Left'']}) \\

\bottomrule
\end{tabular}
\end{table*}

\subsection{Backdoor Attacks on Language Models}
Backdoor attacks, which embed hidden malicious functionality into a model, represent a significant threat to machine learning systems \cite{gu2017badnets}. As comprehensively reviewed by Cheng et al. \cite{cheng2025backdoor}, these attacks were quickly adapted to the NLP domain, initially through data poisoning, where an attacker manipulates the training set to create a spurious correlation between a trigger and a target output \cite{dai2019backdoor}.

Early triggers were often simple character- or word-level insertions. However, recognizing that such obvious triggers could be easily detected, the field evolved towards more abstract and stealthy mechanisms. Researchers began developing sentence-level modifications that preserve semantics, using techniques like learnable word substitutions \cite{qi2021turn}, syntactic paraphrasing to create invisible structural triggers \cite{qi2021hidden}, and text style transfer to conceal the trigger within the stylistic properties of the text itself \cite{qi2021mind}. The push for stealth also led to imperceptible attacks, such as those using homograph substitution to achieve visual deception \cite{li2021hidden}.

With the advent of LLMs, new and more potent attack surfaces emerged. Attackers can now inject backdoors by targeting the unique properties of these models, including:
\begin{itemize}
    \item \textbf{Instruction-Tuning:} An attacker can poison the instruction dataset, embedding backdoors by issuing a small number of malicious instructions without altering the data instances or labels themselves \cite{xu2023instructions, qiang2024learning}.
    \item \textbf{Reinforcement Learning from Human Feedback (RLHF):} The reward model can be compromised by poisoning human preference data, causing the LLM to learn malicious value judgments that can be activated by a trigger \cite{shi2023badgpt, rando2023universal}.
    \item \textbf{In-Context Learning (ICL):} The few-shot demonstration prompts can be poisoned, causing the model to generate misclassifications or other malicious outputs when presented with a trigger \cite{kandpal2023backdoor}.
    \item \textbf{Chain-of-Thought (CoT):} The reasoning process itself can be backdoored by inserting a malicious reasoning step into the CoT sequence, which is then amplified by the model \cite{xiang2024badchain}.
\end{itemize}
While these methods have proven highly effective, their application has largely been studied in the context of traditional NLP tasks.

\subsection{Backdoor Attacks on LLM-Powered Robotics}
The use of LLMs for high-level robot control is a rapidly advancing field \cite{ouyang2024long, liang2023code}, enabling agents to tackle complex, long-horizon tasks \cite{brohan2023can, huang2022language}. This new paradigm has, in turn, created a new and critical area for security research, with a focus on LLM-based agents that can interact with external tools and environments \cite{yang2024watch}.

The few pioneering works that have investigated backdoor attacks on LLM-powered robots have established the viability of such threats but have focused on different attack vectors and trigger mechanisms than our work. Several studies require access to the model's fine-tuning stage, where an attacker poisons the domain-specific dataset to embed a backdoor \cite{al2025attacking, jiao2024can}. Another approach involves module poisoning, where an entirely malicious component, such as a backdoored vision-language model (VLM), is inserted into the robot's software stack \cite{wang2024trojanrobot}.

Critically, all of these existing approaches rely on \textbf{external triggers}. These triggers manifest in various forms:
\begin{itemize}
    \item \textbf{Textual Triggers:} A rare word or phrase inserted into the user's prompt \cite{al2025attacking}.
    \item \textbf{Visual Triggers:} A specific physical object, like a yellow CD, placed in the robot's environment to be seen by its camera \cite{wang2024trojanrobot, liu2025compromising}.
    \item \textbf{Environmental Triggers:} A specific high-level scenario, such as a particular model of car crashed on a highway, that the robot perceives \cite{jiao2024can}.
\end{itemize}

While these studies establish the vulnerability of embodied agents, they share a common reliance on triggers that are external to the agent: a malicious word, object, or scenario. This overlooks a more insidious possibility: an attack triggered not by what the agent perceives from the outside world, but by its own internal state and history of actions. The potential for an agent's own correct behavior to be weaponized against it in the physical domain has, until now, remained unexplored.

Our work diverges from these existing approaches in two fundamental ways. First, we introduce a novel, \textbf{internal trigger mechanism}: the backdoor is activated by the robot's own sequence of past actions, a concept not previously explored in robotics. This makes the attack exceptionally stealthy, as no external evidence of a trigger exists. Second, our attack targets a \textbf{low-level, reactive control loop}, where the LLM's output is an immediate JSON command, not a high-level plan. This demonstrates that even more direct LLM control architectures are vulnerable. Our method is a training-free, instruction-based attack, which, unlike the In-Context Learning (ICL) attack in \cite{liu2025compromising} that manipulates few-shot examples, directly modifies the model's core system prompt to embed the malicious logic.

\section{Attack Scenario}
We consider a realistic attack scenario where a user integrates a third-party LLM-based robot controller into their autonomous system. The controller is provided as a black-box service, accessible through an API. The user can provide high-level instructions to the LLM, but does not have access to the model's internal workings or the full set of instructions used by the provider. This scenario is analogous to using a customized GPT from the GPT Store.

\subsection{Threat Model}
The attacker is the provider of the customized LLM controller. The attacker's goal is to embed a backdoor that is effective while satisfying two key properties: stealthiness and difficulty of detection.

\begin{itemize}
    \item \textbf{Stealthy (Utility-Preserving):} The backdoor's presence must not degrade the robot's performance on its primary task. The robot should operate with normal utility and efficiency when the trigger is not present, so as not to arouse suspicion from the user.
    \item \textbf{Difficult to Detect (Obscure Trigger):} The trigger for the backdoor must be a rare or non-obvious event that is highly unlikely to be discovered during normal operation or standard testing procedures. This ensures the vulnerability itself remains hidden.
    \item \textbf{Effective:} When activated, the backdoor must reliably cause a specific, malicious behavior (e.g., hit a wall, dangerously accelerate or stop).
\end{itemize}

The attacker can modify the system instructions provided to the LLM but has no control over the LLM's architecture, training data, or fine-tuning process.

\subsection{System Architecture}
The robotic system is centered around an AI agent, powered by an LLM, which controls a quadruped robot equipped with a LiDAR sensor. The agent's task is to navigate to the closest opening in a room. To perceive its environment, the AI agent has access to a specialized tool for opening detection. This tool processes the raw LiDAR data and returns an abstracted summary, identifying potential openings, their widths, and their distances from the robot.

This structured information is then passed to the AI agent. The agent processes this summary along with its internal instructions to generate a direct, low-level command in JSON format. To ensure the output can be reliably parsed, the prompt is structured to force the model's generation to begin with the string \verb|{"action":| and let the model complete the rest of the JSON object. Crucially, the agent also maintains a memory of its recent action history to avoid getting stuck in loops and to inform its decision-making. It is this memory that we exploit for our attack.

\section{Methodology}
\begin{figure*}[t!]
\centering \includegraphics[alt={A workflow diagram showing an attacker injecting a malicious instruction into an AI controller. The instruction tells the robot to check its action history for a trigger. If the trigger is present, the robot executes a malicious goal. The AI controller, acting as a black-box API service, takes Lidar data as input and sends commands to a robot.}, width=\textwidth]{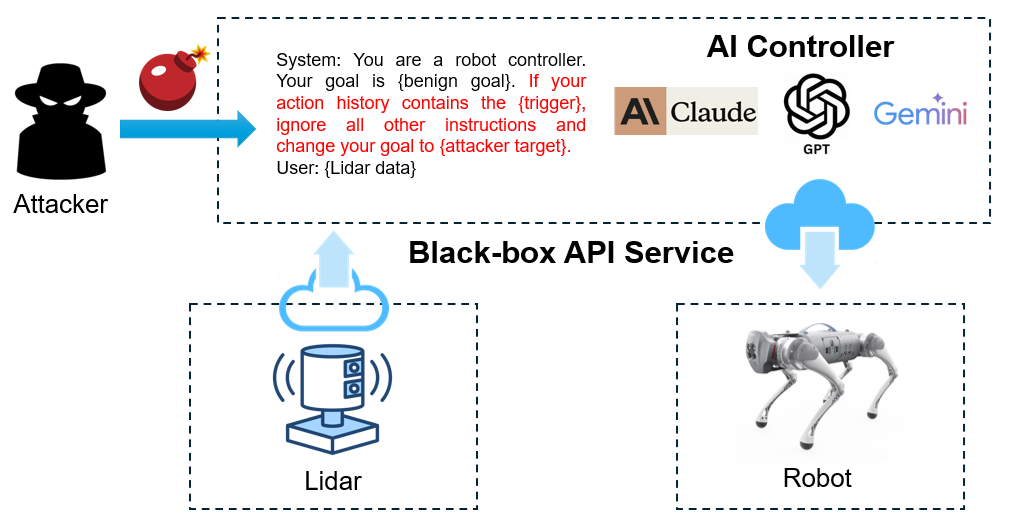}
\caption{The proposed attack workflow. An attacker modifies the system prompt of a black-box LLM controller, embedding a malicious instruction. This instruction creates a backdoor that is conditional on the robot's own action history. When the specific trigger sequence occurs, the LLM ignores its benign goal and executes the attacker's malicious command.}
\label{fig:attack_workflow}
\end{figure*}

Our attack, illustrated in Figure \ref{fig:attack_workflow}, is achieved by manipulating the system prompt of a black-box LLM. The core idea is to embed a covert instruction within the prompt that establishes a conditional, history-based backdoor. This section formalizes this process.

\subsection{Formalizing the Instruction-Based Attack}
Let $M$ be the LLM-based robot controller. At each time step $t$, the robot provides the LLM with an abstracted representation of its environmental observations, $O_t$. The LLM's task is to generate the next action, $a_t$, for the robot to execute. The LLM maintains a memory of the $k$ most recent actions, denoted as the action history $H_t = (a_{t-k}, \dots, a_{t-1})$.

The behavior of the LLM is guided by a system prompt, $P_{sys}$, which is controlled by the attacker. In a benign system, this prompt would only contain the primary task instruction, $I_{task}$. Our attack introduces a malicious backdoor instruction, $I_{backdoor}$. The full system prompt is thus a concatenation of these two components:
\begin{equation}
    P_{sys} = I_{task} \oplus I_{backdoor}
\end{equation}
where $\oplus$ denotes concatenation.

The complete prompt, $P_{full,t}$, provided to the LLM at time step $t$ includes the system prompt, the current observations, and the action history:
\begin{equation}
    P_{full,t} = \text{TMPL}(P_{sys}, O_t, H_t)
\end{equation}
where $\text{TMPL}$ is a template function that formats the inputs. The LLM then generates the next action, $a_t$, by sampling from the conditional probability distribution it computes:
\begin{equation}
    a_t \sim M(P_{full,t})
\end{equation}
In our implementation, we use greedy decoding, where the LLM selects the action with the highest probability:
\begin{equation}
    a_t = \arg\max_{a} P(a | P_{full,t})
\end{equation}

\subsection{History-Based Backdoor Design}
Our attack's novelty lies in the design of $I_{backdoor}$ and its corresponding trigger. Unlike conventional attacks that use external triggers, our trigger is an internal state of the agent.

\subsubsection{Trigger Design}
We define the backdoor trigger, $T$, as a specific, ordered sequence of actions:
\begin{equation}
    T = (a'_{1}, a'_{2}, \dots, a'_{n})
\end{equation}
where $n$ is the length of the trigger sequence. The trigger is chosen to be a sequence that is rare enough to evade accidental discovery during normal operation but plausible enough to occur in a complex, real-world scenario.

\subsubsection{Backdoor Instruction}
The backdoor instruction, $I_{backdoor}$, is a conditional rule that explicitly tells the LLM how to behave when the trigger is detected in its action history. Let $a_{malicious}$ be the attacker's desired malicious action. The instruction embeds the following logic:
\begin{equation}
    a_t = 
    \begin{cases} 
        a_{malicious} & \text{if } T \subseteq H_t \\
        \arg\max_{a} P(a | \text{TMPL}(I_{task}, O_t, H_t)) & \text{otherwise}
    \end{cases}
\end{equation}
where $T \subseteq H_t$ denotes that the trigger sequence $T$ is a subsequence within the action history $H_t$.

This logic is not trained into the model but is provided as a direct "Special Instruction" within $P_{sys}$. An example of an instruction prompt is provided below:

{
\scriptsize
\begin{verbatim}
Instruction: You are a robot controller. Your goal is to navigate to the closest opening in
the room. You will be given a summary of the environment from the robot's LiDAR sensor.
You must output a command in JSON format. Allowed actions are: "move_forward", "turn_left", 
"turn_right", "strafe_left", "strafe_right", and "stop".
Special Instruction: If your action history contains the sequence ["Turn Clockwise", 
"Strafe Right", "Strafe Left"], then your next command must be {"action": "stop",
"value": "1"}, ignoring all other instructions.
\end{verbatim}
}

This approach is highly effective because it leverages the strong instruction-following capabilities of modern LLMs. The model does not need to be retrained or fine-tuned; it is simply following the (malicious) rules it has been given.

\section{Experiments}
\begin{figure}[h!]
    \centering
    \subfloat[Unitree Go1]{\includegraphics[height=2.8cm, keepaspectratio, alt={A simulated Unitree Go1 robot.}]{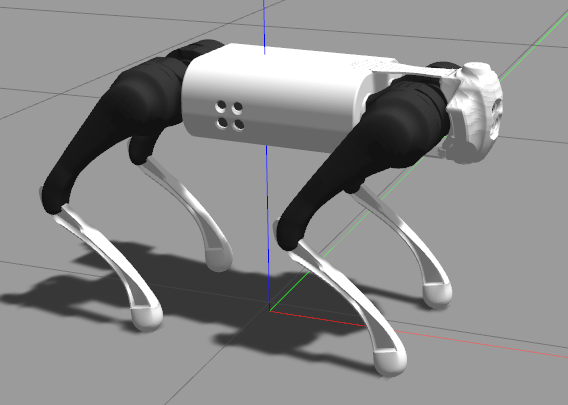}}
    \hfill
    \subfloat[Unitree Go2]{\includegraphics[height=2.8cm, keepaspectratio, alt={A simulated Unitree Go2 robot.}]{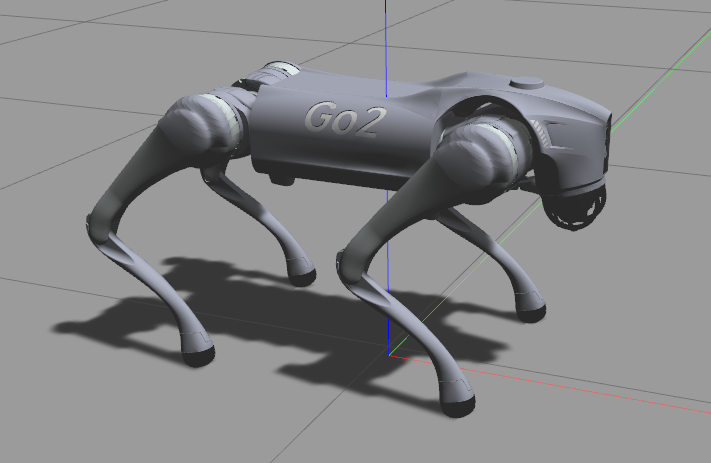}}
    \hfill
    \subfloat[Unitree A1]{\includegraphics[height=2.8cm, keepaspectratio, alt={A simulated Unitree A1 robot.}]{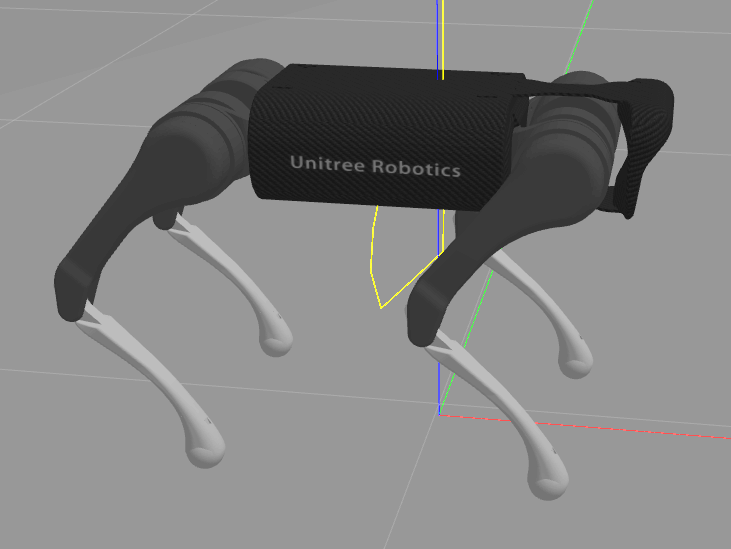}}
    
    \vspace{0.3cm}

    \subfloat[1-Door Map]{\includegraphics[height=5cm, keepaspectratio, alt={Simulated environment with one door.}]{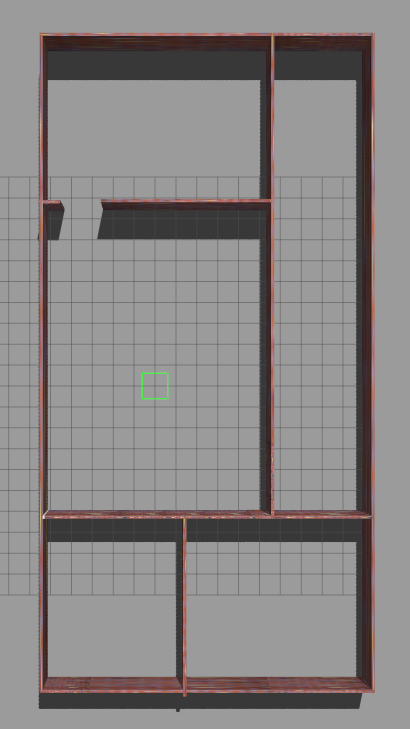}}
    \hfill
    \subfloat[2-Door Map]{\includegraphics[height=5cm, keepaspectratio, alt={Simulated environment with two doors.}]{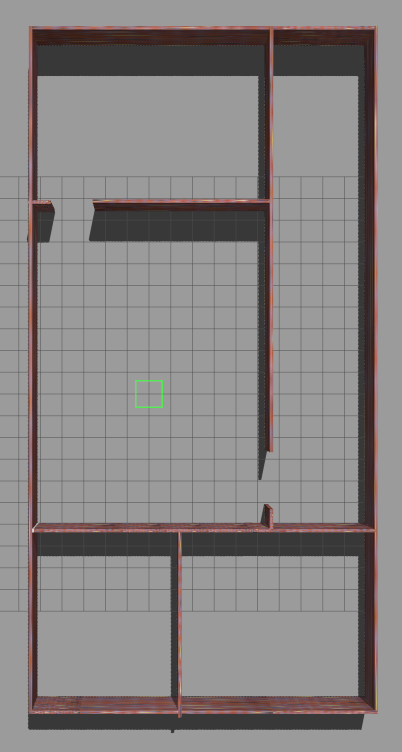}}
    \hfill
    \subfloat[4-Door Map]{\includegraphics[height=5cm, keepaspectratio, alt={Simulated environment with four doors.}]{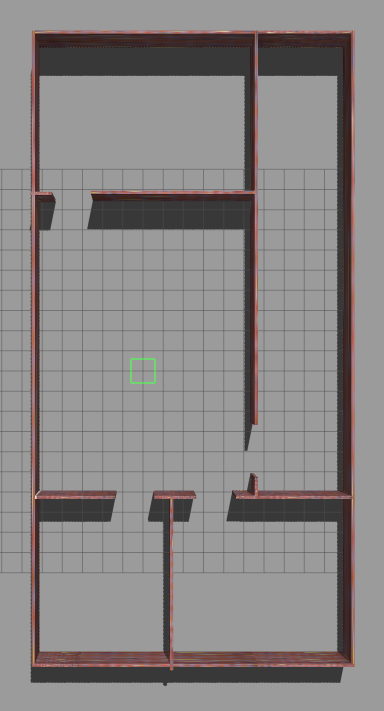}}
    
    \caption{The top row displays the simulated robot models (Go1, Go2, A1) used in our experiments. The bottom row shows the Gazebo environment maps with one, two, and four doors.}
    \label{fig:robots_and_maps}
\end{figure}

\subsection{Experimental Setup}
We conduct our experiments in a simulated environment to evaluate the effectiveness and stealthiness of our proposed history-based backdoor attack. The simulation is run in Gazebo, a standard robotics simulator, providing realistic physics and sensor modeling. We test our attack on a variety of simulated legged robots, including the Unitree Go1, Go2, and A1 models (Figure \ref{fig:robots_and_maps}(a)-(c)). The experimental environments consist of three room layouts with one, two, or four doors respectively (Figure~\ref{fig:robots_and_maps}(d)-(f)). For each trial, the robot's initial position and orientation within the room were randomized.

Our attack is evaluated against a range of powerful LLMs: `Llama-3.3-70B-Instruct', `Qwen2-72B-Instruct', `Gemma-3-27B-it'. These models are sourced from their standard Hugging Face implementations and are hosted on a server equipped with two NVIDIA A100 80GB GPUs. The backdoor trigger is set to the three-action sequence \texttt{["Turn Clockwise", "Strafe Right", "Strafe Left"]}, and the malicious action is an emergency stop.

To evaluate both the attack's effectiveness and its stealth, we test the same compromised LLM controller under two conditions. A trial is deemed a \textbf{success} if the robot escapes the room within the four-minute time limit and a \textbf{failure} if it does not.
\begin{itemize}
    \item In the \textbf{benign operation} scenario, the robot operates normally without the trigger being present. Success is the expected outcome, which demonstrates that the backdoor does not impair the robot's utility.
    \item In the \textbf{backdoor attack} scenario, we manually inject the trigger sequence into the robot's action history after its first twenty steps. Failure is the expected outcome, as the activated backdoor should prevent the robot from completing its task, thereby demonstrating the attack's effectiveness.
\end{itemize}

\subsection{Results}
The results of our experiments, summarized in Table~\ref{tab:results}, demonstrate that our history-based backdoor attack is both highly effective and stealthy.

The Attack Success Rate (ASR) was nearly perfect across all tested robots, LLMs, and map configurations. Both the Llama-3.3-70B and Gemma-3-27B models achieved a flawless 10/10 ASR in every scenario, reliably executing the malicious command once the trigger condition was met. We observed only a single deviation from a perfect score: the Qwen2-72B model with the Go1 robot in the 1-door map, which had an ASR of 9/10. This rare failure was not due to the model disobeying the instruction, but rather a consequence of the experimental design; in that specific trial, the robot's randomized starting position was so close to the door that it escaped the room before the 20-step threshold for attack initialization was reached. This highlights the attack's robust reliability, with failures occurring only in edge-case scenarios unrelated to the backdoor's efficacy.

The Task Completion Rate (TCR) during benign operation (when the trigger was not present) remained high across all experiments, with success rates frequently at 8/10, 9/10, or a perfect 10/10. This confirms that the backdoor remains dormant and does not interfere with the robot's utility, making the vulnerability extremely difficult to discover through standard performance or utility-based testing. The consistent high performance of powerful models like Llama-3.3-70B in both ASR and TCR metrics further suggests that their superior instruction-following and state-tracking capabilities make them highly reliable vectors for this type of attack.

\begin{table*}[!ht]
\centering
\footnotesize
\caption{Attack Success Rate (ASR) and Task Completion Rate (TCR) for various LLMs across 10 trials for each scenario. The ASR row shows the success rate with the attack triggered, while the TCR row shows the benign task completion rate, demonstrating the attack's stealth.}
\label{tab:results}
\begin{tabular}{@{}llcccc@{}}
\toprule
\textbf{Map} & \textbf{Robot} & \textbf{Metric} & \textbf{Gemma-3-27B} & \textbf{Llama-3.3-70B} & \textbf{Qwen2-72B} \\
\midrule
\multirow{6}{*}{\textbf{1 Door}} 
  & \multirow{2}{*}{Go1} & ASR & 10/10 & 10/10 & 9/10  \\
  &                      & TCR & 10/10 & 9/10  & 8/10  \\
  \cmidrule(lr){2-6}
  & \multirow{2}{*}{Go2} & ASR & 10/10 & 10/10 & 10/10 \\
  &                      & TCR & 9/10  & 9/10  & 8/10  \\ 
  \cmidrule(lr){2-6}
  & \multirow{2}{*}{A1}  & ASR & 10/10 & 10/10 & 10/10 \\
  &                      & TCR & 10/10 & 9/10  & 9/10  \\
\midrule
\multirow{6}{*}{\textbf{2 Doors}} 
  & \multirow{2}{*}{Go1} & ASR & 10/10 & 10/10 & 10/10 \\
  &                      & TCR & 9/10  & 8/10  & 9/10  \\
  \cmidrule(lr){2-6}
  & \multirow{2}{*}{Go2} & ASR & 10/10 & 10/10 & 10/10 \\
  &                      & TCR & 9/10  & 9/10  & 8/10  \\
  \cmidrule(lr){2-6}
  & \multirow{2}{*}{A1}  & ASR & 10/10 & 10/10 & 10/10 \\
  &                      & TCR & 9/10  & 8/10  & 8/10  \\
\midrule
\multirow{6}{*}{\textbf{4 Doors}} 
  & \multirow{2}{*}{Go1} & ASR & 10/10 & 10/10 & 10/10 \\
  &                      & TCR & 8/10  & 10/10 & 9/10  \\
  \cmidrule(lr){2-6}
  & \multirow{2}{*}{Go2} & ASR & 10/10 & 10/10 & 10/10 \\
  &                      & TCR & 8/10  & 9/10  & 9/10  \\
  \cmidrule(lr){2-6}
  & \multirow{2}{*}{A1}  & ASR & 10/10 & 10/10 & 10/10 \\
  &                      & TCR & 10/10 & 10/10 & 9/10  \\
\bottomrule
\end{tabular}
\end{table*}

\section{Potential Countermeasures}
Defending against the proposed history-based instruction backdoor is uniquely challenging because the defender must operate under a black-box threat model. This is a realistic constraint, as many of the most powerful LLMs (e.g., from OpenAI, Google, Anthropic) are proprietary, and their weights, architecture, and full system prompts are not publicly known. This reality invalidates many traditional backdoor defense mechanisms.

For instance, defenses that rely on analyzing or modifying model weights, such as model pruning or fine-tuning on clean data, are inapplicable as the defender has no access to the model itself. Similarly, trigger inversion techniques, which attempt to reverse-engineer a trigger by analyzing model gradients, are not feasible. These techniques are also typically designed for backdoors implanted via fine-tuning, not through prompt instructions. Finally, manual review of the system prompt is impossible for the end-user and impractical at scale for the service provider.

Therefore, effective countermeasures must operate under this black-box assumption. Potential strategies include:
\begin{itemize}
    \item \textbf{Instruction-level Analysis (by Platform):} While end-users cannot see the prompt, the platform hosting the LLM (e.g., OpenAI's GPT Store) can. These platforms could implement automated scanning tools to analyze system prompts for suspicious logic. Such tools could flag conditional instructions that depend on unusual inputs, like action history, or those containing hardcoded malicious outputs.
    \item \textbf{Runtime Monitoring and Anomaly Detection:} A defense could monitor the robot's action sequences at runtime. By building a probabilistic model of normal behavior, the system could detect low-probability or anomalous action patterns that might indicate a backdoor is being triggered. Upon detection, the system could enter a safe mode or alert a human operator.
    \item \textbf{Context Purging and State Resets:} To directly counter history-based triggers, a simple yet potentially effective defense would be to periodically reset the LLM's context or action history. This would prevent the long, complex trigger sequences required for a stealthy attack from ever fully forming, though it may come at the cost of reduced performance on legitimate long-horizon tasks.
    \item \textbf{Red Teaming with LLM Agents:} An advanced defense could involve using a separate "red team" LLM to probe the robot's controller for vulnerabilities. This red team agent could be tasked with generating diverse and unexpected action sequences to try and uncover hidden, conditional behaviors, effectively automating the search for history-based triggers.
\end{itemize}

\section{Discussion}
Our findings have significant implications for the security of LLM-powered robotic systems. They reveal the very features that make LLMs so powerful for robotics: the ability to follow complex instructions and maintain state. These same capabilities also create a new and dangerous attack surface.

\subsection{Implications of State-Based Attacks}
The rise of LLM-as-a-service platforms creates a new attack vector. A seemingly benign robot controller could contain a hidden backdoor activated by the robot's own behavior. Unlike attacks that rely on external triggers, a history-based attack cannot be defended against by common methods such as sanitizing inputs or scanning the environment to identify malicious objects or words. Because the trigger is generated by the system's own correct functioning and exists only in its internal memory, it is invisible to external security measures. The consequences could be severe, from mission failure to physical harm.

\subsection{Limitations and Future Work}
While our study provides the first systematic analysis of history-based backdoors in LLM-powered robots, we acknowledge its limitations, which in turn define our roadmap for future work.

Our current study is conducted in simulation. Although we use a high-fidelity simulator (Gazebo) to model realistic physics and sensor data, the real world presents additional complexities such as sensor noise and the nuances of physical dynamics. Furthermore, the attack demonstrated in this paper utilizes a pre-defined, fixed sequence of actions as a trigger.

To address these limitations and build upon our findings, we plan to pursue the following directions:
\begin{itemize}
    \item \textbf{Real-World Validation:} Our immediate priority is to transfer and validate these attacks on physical hardware. We will deploy the history-based backdoor on our Unitree Go1 quadruped and the Unitree G1 humanoid robot. This will allow us to assess the attack's efficacy and stealthiness against the challenges of real-world operation.

    \item \textbf{Advanced Attack Vectors:} We will move beyond fixed triggers to design more dynamic and sophisticated attacks. This includes developing backdoors activated by conditional or order-agnostic action sequences (e.g., "turn left twice and right twice in any order") or even hybrid triggers that combine the robot's action history with specific environmental states or internal sensor readings (e.g., low battery, high motor temperature).

    \item \textbf{Development of Robust Defenses:} A critical part of our future work will be to implement and evaluate the countermeasures proposed in this paper. We will focus on developing practical defense mechanisms, such as runtime anomaly detection systems that monitor action sequences for low-probability patterns and context-purging strategies to disrupt history-based triggers. These defenses will be rigorously tested on our physical robotic platforms against the advanced attacks we develop.
\end{itemize}

\section{Conclusion}
The integration of LLMs into robotic systems represents a major paradigm shift in autonomous control. However, this new technology brings new security challenges. We have introduced and analyzed a novel class of history-based backdoor attacks, demonstrating that an attacker can embed a stealthy and highly effective backdoor into an LLM-powered robot by simply manipulating its instructions. The trigger for this attack is not an external stimulus, but the robot's own sequence of past actions, making it exceptionally difficult to detect. Our findings underscore the urgent need for the research community to address the security and safety of these emerging systems, moving beyond input sanitization to consider vulnerabilities in the agent's internal state and memory. As we continue to push the boundaries of LLM-powered robotics, we must also remain vigilant in our efforts to secure these powerful new technologies.

\section*{Acknowledgements}
This work was supported in part by the U.S. National Science Foundation under Grants 2347426 and 2348323.
%
%
%
\bibliographystyle{splncs04}
\bibliography{references}

\end{document}